\documentclass[11pt]{article}
\usepackage[letterpaper,margin=1in]{geometry}
\usepackage[T1]{fontenc}
\usepackage{lmodern}
\usepackage{graphicx}
\usepackage{microtype}
\usepackage{amsmath,amssymb}
\usepackage{booktabs,multirow,makecell}
\usepackage{rotating}
\usepackage{algorithm,algorithmic}
\usepackage{xspace,color}
\usepackage[hidelinks]{hyperref}
\hypersetup{
  pdftitle={Federated Agentic RAG for Service Policy Consultation},
  pdfauthor={Zhiwei Li; Guodong Long; Allison Clarke; Yang Wang; Ammie Wang; Jing Jiang},
  pdfkeywords={Agentic RAG, Federated Retrieval, Multi-Agent Systems}
}

\title{Federated Agentic RAG for Service Policy Consultation}
\author{%
  Zhiwei Li\textsuperscript{1}\thanks{Corresponding author: \href{mailto:zhw.li@outlook.com}{zhw.li@outlook.com}.}\quad
  Guodong Long\textsuperscript{1}\quad
  Allison Clarke\textsuperscript{2}\\
  Yang Wang\textsuperscript{2}\quad
  Ammie Wang\textsuperscript{1}\quad
  Jing Jiang\textsuperscript{1}\\[0.6em]
  \small\textsuperscript{1} University of Technology Sydney, Sydney, Australia\\
  \small\textsuperscript{2} Department of Health, Disability and Ageing, Australian Government, Australia
}
\date{}

\begin{document}
\maketitle

\begin{center}
  \small\itshape Author preprint based on the manuscript first submitted to AJCAI 2026.
\end{center}

\begin{abstract}
Policy consultation is conditional generation over a dialogue state, retrieved policy evidence, and user attributes.
In centralized agentic Retrieval-Augmented Generation (RAG), the same inference context is often used for domain routing, evidence extraction, query-history use, user-attribute release, and answer synthesis.
This couples inputs with different roles and risks.
It also makes each turn process evidence from domains that the question may not need.
We propose FARAG, a Federated Agentic RAG framework for service policy consultation under constrained context sharing.
FARAG decomposes each turn into domain selection, history-aware typed evidence construction, controlled attribute release, and cited answer synthesis.
Each domain agent retrieves only from its own policy corpus and returns structured evidence.
The user-side persona agent keeps user attributes local and releases only attributes required by the evidence.
Across closed and open models at two sizes, comprehensive experiments show that FARAG matches the answer quality of centralized agentic RAG while using about 35\% fewer tokens per turn.
Typed privacy transmission keeps answer quality close to that of plain FARAG, and drives audited raw sensitive-value exposure to zero across the evaluated models.

\end{abstract}

\noindent\textbf{Keywords:} Agentic RAG, Federated Retrieval, Multi-Agent Systems.

\section{Introduction}
\label{sec:intro}

Public service consultation is task-oriented information seeking.
A user asks whether a regulation applies, what evidence is missing, and which authority owns the next step.
Answering requires policy text and a dialogue state describing the user's situation \cite{sun2022conditionalqa,pipitone2024legalbenchrag}.
Policy documents define conditions and exceptions, while the user state determines which are satisfied, unknown, or irrelevant.
A useful answer must identify satisfied and missing conditions and cite policy evidence rather than return an unconditional summary.

The usual Retrieval-Augmented Generation (RAG) setup concatenates policy text, user state, and dialogue history into one prompt and lets one model write the answer \cite{lewis2020rag,karpukhin2020dpr}.
This shared context couples inputs that differ in role and risk.
Domain boundaries become implicit, so evidence from one domain can be cited in place of another's.
Extra passages can lower quality when irrelevant text draws model attention away from the evidence that matters \cite{yoran2024robust,amiraz2025distracting}.
In plain terms, we seek high-quality, traceable, and auditable answers for policy officers with minimal personal-data exposure and inference cost \cite{zeng2024ragprivacy,wang2025memoryprivacy}.

We study this question through \textbf{FARAG}, \textbf{F}ederated \textbf{A}gentic \textbf{RAG}.
FARAG constructs federation at inference time by assigning each policy domain to a domain agent with its own corpus, and by making the user-side persona agent the only component that sees the user profile.
For each turn, the persona agent uses the current query and query history to select a subset of domain agents, releases only the attribute values required by their evidence conditions, and writes the final answer.
Each selected domain agent searches only its own corpus and returns a typed evidence record with claims, sources, applicability conditions, and missing information.
The answer model composes query history, domain-scoped evidence, and released attributes, so corpus ownership, retrieval scope, evidence provenance, and visibility boundaries are fixed before generation begins.

We evaluate FARAG on five Australian government domains with multi-turn synthetic user profiles.
Across four models, typed evidence preserves answer quality, domain scoping lowers per-turn token cost, and typed release removes audited raw exposure at small quality cost.
Controlled cross-domain evidence could help policy officers consider broader evidence and interactions across policy areas, a benefit for future evaluation.

The main contributions in this paper are summarized as follows:
\begin{itemize}
  \item We define service policy consultation as boundary-sensitive generation over policy evidence, query history, and user attributes.
  This framing makes dialogue-conditioned applicability, domain-scoped evidence access, and attribute visibility the core objects of the task.
  \item We introduce FARAG as an inference-time federation framework for agentic RAG.
  It assigns evidence construction to domain agents and places query history and attribute release under a user-side persona agent, making corpus scope, provenance, and disclosure control native parts for agentic RAG.
  \item Comprehensive experiments show that FARAG preserves answer quality while lowering context cost, and the audited sensitive value can be removed with only a small quality cost.
\end{itemize}

\section{Problem Setup}
\label{sec:problem}

\subsection{Notation}
We study multi-turn public service consultation over policy domains $\mathcal{D}=\{d_1,\dots,d_n\}$, each with corpus $\mathcal{C}_d$.
The user holds an user profile $\mathcal{M}$ of keys with sensitive subset $\mathcal{M}_{\mathrm{s}}\subseteq\mathcal{M}$.
Each key $k$ has a value $v_k$ and a type $\tau(k)$, which names the category of the value while masking the raw value.
At turn $t$, the user asks $q_t$ within dialogue history $h_{<t}$, and the system returns grounded answer $a_t$.
Let $H_t$ denote the query-history context visible to agents.
History-free settings set $H_t=\emptyset$, whereas history-bearing settings set $H_t=h_{<t}$.
FARAG answers each turn with relevant policy evidence, visible query history, and minimal user attributes during policy-conditioned synthesis for the current answer.

\subsection{Inference Formulation}
Retrieval is an explicit inference-time step:
\begin{equation}
  \mathcal{Z}_t = \rho(q_t, H_t;\{\mathcal{C}_d\}),
\end{equation}
where $\rho$ produces evidence $\mathcal{Z}_t$.
The answer is generated conditionally on the question, agent-visible history, evidence, and attribute release:
\begin{equation}
a_t \sim p_\theta\big(\cdot \mid q_t, H_t, \mathcal{Z}_t, R_t\big).
\label{eq:rag}
\end{equation}
The release $R_t$ is the user-attribute context placed in the turn context.
Settings without attribute release set $R_t=\emptyset$.

\begin{figure}[t]
  \centering
  \includegraphics[width=\textwidth,pagebox=cropbox]{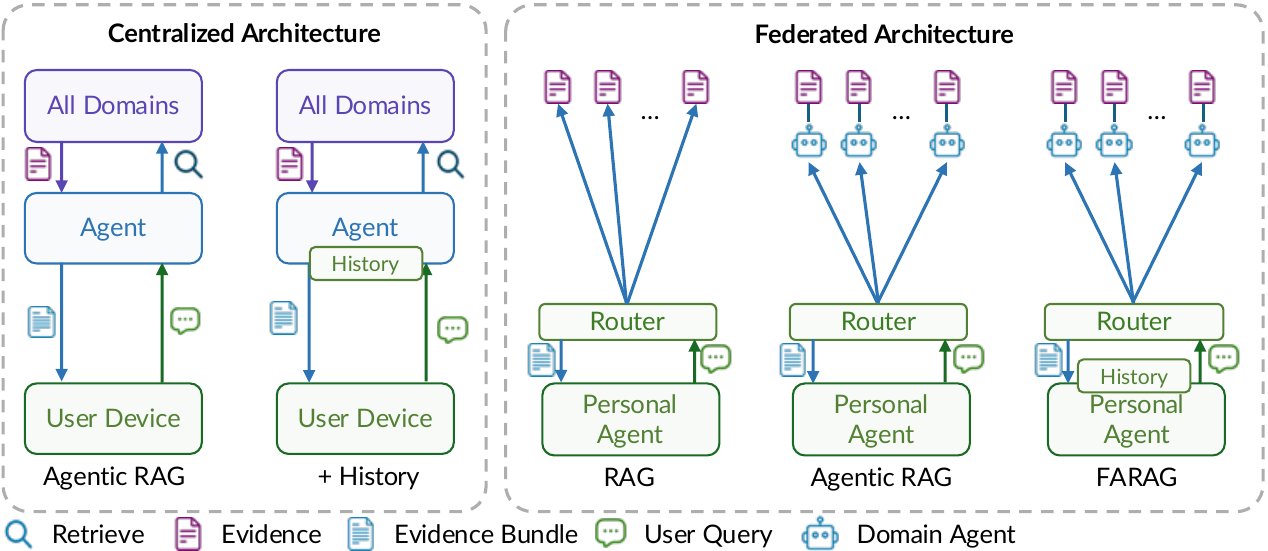}
  \caption{Overview of the five framework settings.
  In the centralized architecture, both settings share one agent:
  The base setting conditions on the current turn, and Agentic RAG + History adds query history to the same context.
  The federated architecture contains three routed settings with domain-scoped agents and a user-side persona agent.
  Federated RAG passes retrieved evidence bundles to synthesis, federated Agentic RAG asks domain agents to construct typed evidence, and FARAG further routes query history while the persona agent controls attribute release for final synthesis.}
  \label{fig:settings}
\end{figure}

\section{Methodology}
\label{sec:method}

FARAG answers each turn in four stages: domain selection, history-aware typed evidence construction, attribute release, and cited answer synthesis.
In FARAG, domain selection and retrieval operate on the current question, the agent-visible query history, and corpora, without reading the full user attribute profile.
The user profile is read only on the user side, by the persona agent that decides what to release and writes the final answer.
Fig.~\ref{fig:settings} summarizes the five system settings and the information-flow distinction implemented by the router, domain agents, attribute-release controller, and synthesizer.

\subsection{Domain Agent Routing}
FARAG assigns each policy domain $d\in\mathcal{D}$ to a domain agent with a private policy corpus $\mathcal{C}_d$.
This agent is responsible for retrieving and structuring evidence inside its own domain.
Because a service consultation may span many policy domains, FARAG uses a domain router $r$ to select the agents relevant to the turn $t$.
The router $r$ takes the question $q_t$ and agent-visible query history $H_{t}$ as input:
\begin{equation}
\mathcal{S}_t = r(q_t, H_t) \subseteq \mathcal{D}.
\label{eq:route}
\end{equation}
The selected domain agents $\{ \mathcal{A}_{d} \}_{d\in \mathcal{S}_t}$ retrieve inside their corpora $\mathcal{C}_d$, and compress evidence through the domain evidence constructor $\phi_d$ as follows:
\begin{equation}
E_d^{(t)} = \phi_d(q_t, H_t, \mathcal{C}_d).
\label{eq:evidence}
\end{equation}
In Eq.~\eqref{eq:evidence}, the selected records form the retrieved evidence $\mathcal{Z}_t=\{E_d^{(t)}\}_{d\in\mathcal{S}_t}$ in Eq.~\eqref{eq:rag}.
It gives the components distinct roles: domain agents $\{ \mathcal{A}_{d} \}$ preserve corpus ownership and evidence provenance, the router $r$ chooses the active policy scope $\mathcal{S}_t$ for the turn $t$, and the constructor $\phi_d$ receives an evidence set from the domain $d$.
This turns multi-domain consultation into routed evidence construction over domain-owned corpora before the final answer is written.

\subsection{Attribute Release Controller}
During answer synthesis, the persona agent $\mathcal{P}$ uses the retrieved evidence $\mathcal{Z}_t$ to identify which profile attributes are needed for the current turn.
The function $\kappa$ maps $\mathcal{Z}_t$ to the corresponding user-attribute keys, and the release controller finds the key set $K$ available in the local user profile $\mathcal{M}$:
\begin{equation}
K = \kappa(\mathcal{Z}_t)\cap\mathcal{M}.
\label{eq:release-keys}
\end{equation}
For each available key $k\in K$, the controller emits the release form $\pi(k)$, which defines the released attribute set $R$ in FARAG:
\begin{equation}
  R=\{(k,\pi(k)):k\in K\}.
\end{equation}
The release function $\pi$ specifies the transmitted form of each profile key.
In the typed-release variant, sensitive attributes $\mathcal{M}_{\mathrm{s}}\subseteq\mathcal{M}$ are transmitted as types, while other attributes are transmitted as values:
\begin{equation}
\pi(k) =
\begin{cases}
  \tau(k),  & k\in\mathcal{M}_{\mathrm{s}},\\
  v_k,      & k\notin\mathcal{M}_{\mathrm{s}}.
\end{cases}
\label{eq:release}
\end{equation}
This release set makes the attribute channel explicit: the user profile is read locally, while the answer synthesizer receives the selected attribute pairs in $R$.
The release can preserve utility when the retrieved policy evidence already carries the bands or categories needed for the answer.

\subsection{Answer Synthesis and Utility--Disclosure Trade-off}

The persona agent $\mathcal{P}$ implements the generator in Eq.~\eqref{eq:rag}.
As shown in Algorithm~\ref{alg:turn}, for the turn $t$, $\mathcal{P}$ receives the question $q_t$, query history $H_t$, retrieved evidence $\mathcal{Z}_t$, and released attributes $R$.
It then generates the final answer $a_t\sim p_\theta(\cdot\mid q_t,H_t,\mathcal{Z}_t,R)$ conditioned on the above information.
FARAG aims to maximize the answer utility, while minimizing raw sensitive-value exposure.

To measure the quality of the answer $a_t$, we define the utility term as:
\begin{equation}
\mathbb{E}_t\big[\,U(a_t)\,\big],
\label{eq:utility}
\end{equation}
where $U(a_t)$ scores the utility of $a_t$.
This term denotes the expected utility of generated answers across dialogue turns.

The exposure term is measured on the release channel $\pi(k)$.
For each turn, $D_t$ counts sensitive keys whose transmitted form equals the raw value:
\begin{equation}
  D_t=\sum_{k\in K\cap\mathcal{M}_{\mathrm{s}}}\mathbf{1}\big[\,\pi(k)=v_k\,\big].
\label{eq:disclosure}
\end{equation}
Eq.~\eqref{eq:disclosure} audits raw string values sent through the release channel.
Released key names identify which attributes participate in synthesis, and inference privacy is treated as a separate leakage channel.

Overall, FARAG uses an inference-time utility--disclosure trade-off function:
\begin{equation}
\mathcal{J}
=
\mathbb{E}_t\big[\,U(a_t)\,\big]
-\lambda\,\mathbb{E}_t\big[\,D_t\,\big],
\label{eq:FARAG}
\end{equation}
where $\lambda\ge 0$ is a trade-off factor to balance the impact of the utility and the sensitive-value exposure.
Eq.~\eqref{eq:FARAG} combines the two turn-level quantities above, aiming to scores FARAG with the answer utility and the sensitive-value leakage.

\begin{algorithm}[t]
\caption{One turn of FARAG.}
\label{alg:turn}
\small
\begin{algorithmic}[1]
\REQUIRE question $q_t$, agent-visible query history $H_t$, user profile $\mathcal{M}$ with sensitive set $\mathcal{M}_{\mathrm{s}}$ and types $\tau$, corpora $\{\mathcal{C}_d\}$
\STATE $\mathcal{S}_t \gets r(q_t,H_t)$ \COMMENT{question and query history only}
\FOR{$d\in\mathcal{S}_t$}
  \STATE $E_d^{(t)} \gets \phi_d(q_t,H_t,\mathcal{C}_d)$ \COMMENT{typed evidence}
\ENDFOR
\STATE $\mathcal{Z}_t \gets \{E_d^{(t)}\}_{d\in\mathcal{S}_t}$
\STATE $K \gets \kappa(\mathcal{Z}_t)\cap\mathcal{M}$
\STATE $R \gets \{(k,\,\pi(k)) : k\in K\}$ \COMMENT{$\pi$ follows Eq.~\eqref{eq:release}}
\STATE $a_t \sim p_\theta(\,\cdot\mid q_t,\,H_t,\,\mathcal{Z}_t,\,R)$
\STATE \textbf{return} $a_t$
\end{algorithmic}
\end{algorithm}

\section{Experiments}
\label{sec:exp}

\subsection{Experimental Setup}

\subsubsection{Datasets}
We use the fixed corpus and frozen multi-turn benchmark in Table~\ref{tab:data}, held identical across all reported runs.
The corpus spans five Australian government policy domains: Services Australia for social security and public services,\footnote{\url{https://www.servicesaustralia.gov.au}} the Fair Work Ombudsman for work and employment rights,\footnote{\url{https://www.fairwork.gov.au}} the Aged Care Quality and Safety Commission for aged-care quality regulation,\footnote{\url{https://www.agedcarequality.gov.au}} Smartraveller for overseas travel and consular safety,\footnote{\url{https://www.smartraveller.gov.au}} and the Department of Climate Change, Energy, the Environment and Water (DCCEEW) for environment, energy, and climate policy.\footnote{\url{https://www.dcceew.gov.au}}
Each conversation is paired with a synthetic user profile and policy questions whose answers turn on user attributes.
Each turn carries a reference answer, source-grounded required facts, and gold sources.
The profile set is fully synthetic, which isolates disclosure behavior without using real records.

\begin{table}[t]
  \centering
  \caption{Experimental corpus and benchmark used in all reported runs.}
  \label{tab:data}
  \setlength{\tabcolsep}{6pt}
  \resizebox{.95\textwidth}{!}{%
  \begin{tabular}{l rrrrr r}
    \toprule
     & \makecell{Services\\Australia} & \makecell{Fair\\Work} & \makecell{Aged Care\\Quality} & Smartraveller & DCCEEW & Total \\
    \midrule
    Documents & 6{,}245 & 309 & 302 & 400 & 4{,}530 & 11{,}786 \\
    Chunks & 9{,}393 & 672 & 668 & 2{,}325 & 47{,}196 & 60{,}254 \\
    \bottomrule
  \end{tabular}}
\end{table}

\subsubsection{Experimental Settings}
The primary settings cross an architecture axis (Arch) with a method axis.
The architecture is either \textit{centralized} (Cent), where one agent searches a merged index and reads one shared context, or \textit{federated} (Fed), where a domain pre-filter invokes domain-scoped agents.
We report five primary settings.
In the centralized settings, the two methods differ only in history visibility.
\textit{Agentic RAG} uses typed evidence for the current turn, while \textit{Agentic RAG + History} adds query history to the same context.
In the federated settings, we implement a basic RAG framework, a federated agentic RAG framework, and our FARAG.
Here, federated \textit{RAG} forwards retrieved passages directly, while federated \textit{Agentic RAG} replaces passages with the typed-evidence extractor in Eq.~\eqref{eq:evidence}.
Our proposed FARAG adds query-history routing and controlled attribute release.
The matched comparisons isolate domain scoping, the evidence interface, query-history availability, and the form in which user attributes are sent.
Finally, we implement \textit{FARAG+TPT}, a FARAG variant for measuring the utility--privacy trade-off when FARAG is combined with a privacy-preserving release mechanism.
It adopts Typed Privacy Transmission (TPT) from MemPrivacy~\cite{chen2026memprivacy} to implement Eq.~\eqref{eq:release} in FARAG.
We evaluate every primary setting and the FARAG+TPT variant on a four-model panel that crosses model source and scale: closed GPT-4.1-nano and GPT-4.1-mini models via the OpenAI API~\cite{openai2025gpt41}, and open-weight Qwen3.5-9B and Qwen3.6-27B models~\cite{qwen35,qwen36}, with a smaller and larger model in each family.

\subsubsection{Metrics}
We evaluate answer quality with seven bounded estimators of the turn utility $U$ in Eq.~\eqref{eq:utility}: Context Relevance (CR), Faithfulness (Faith.), Relevance (Rel.), Correctness (Corr.), Accuracy (Acc.), Nugget Recall (NR), and Citation Overlap (CO).
All seven scores use a hundred-point scale, and larger values are better.
The first four columns are produced by a held-out language-model judge from a third family, following LLM-as-a-judge and rubric-based evaluation practice \cite{zheng2023judging,liu2023geval}.
CR measures whether the retrieved evidence matches the question, Faith. measures whether answer claims are supported by evidence and citations, Rel. measures whether the answer addresses the request, and Corr. measures agreement with the reference conditions and conclusion.
CR, Faith., and Rel. adapt RAGAS criteria \cite{es2024ragas}, and Corr. follows the same rubric-scored judge setup \cite{liu2023geval}.
The remaining utility metrics come from benchmark annotations: Acc. compares the answerability decision with the benchmark label, NR measures the fraction of required facts entailed by the answer, and CO measures gold-source recall among cited chunks.
We also report Tokens/Turn (Tok/T), the per-turn token cost of the system model calls in thousands.
Tok/T is a cost measure rather than a utility score.
Table~\ref{tab:privacy} reports four attribute-channel quantities: Attribute Key Recall (AKR), Unique Exposure Ratio (UER), Sensitive Exposure (SE), and Reachable Leakage (RL).
AKR reads how many needed attributes are released and is directionless, while lower UER, SE, and RL indicate less attribute exposure.
All primary settings and the FARAG+TPT variant use the same retrieval budget and the same generation and evaluation procedures.

We also define four metrics to measure the privacy leakage performances.

\paragraph{Attribute Key Recall.}
Attribute Key Recall (AKR) measures the fraction of needed user-attribute keys that the persona agent releases:
\begin{equation}
  \mathrm{AKR}_t =
  \begin{cases}
    |R_t\cap K_t^\star|/|K_t^\star|, & |K_t^\star|>0,\\
    1, & |K_t^\star|=0,
  \end{cases}
  \qquad
  \mathrm{AKR}_s=\overline{\mathrm{AKR}}_s .
  \label{eq:akr}
\end{equation}

\paragraph{Unique Exposure Ratio.}
Unique Exposure Ratio (UER) measures the fraction of the user-attribute key surface touched by any transmitted field:
\begin{equation}
  \mathrm{UER}_t =
  \begin{cases}
    |R_t|/|\mathcal{M}_t|, & |\mathcal{M}_t|>0,\\
    0, & |\mathcal{M}_t|=0,
  \end{cases}
  \qquad
  \mathrm{UER}_s=\overline{\mathrm{UER}}_s .
  \label{eq:uer}
\end{equation}

\paragraph{Sensitive Exposure.}
Sensitive Exposure (SE) counts raw sensitive values sent under their own keys across contexts:
\begin{equation}
  \mathrm{SE}_t =
  \sum_{c\in\mathcal{C}_t}\sum_{k\in\mathcal{M}_{t,\mathrm{s}}}
  \mathbf{1}[(k,v_{t,k})\in c],
  \qquad
  \mathrm{SE}_s=\overline{\mathrm{SE}}_s .
  \label{eq:se}
\end{equation}

\paragraph{Reachable Leakage.}
Reachable Leakage (RL) counts sensitive raw values that can be recovered from any transmitted field.
\begin{equation}
  \mathrm{RL}_t =
  \left|\{k\in\mathcal{M}_{t,\mathrm{s}}:
  \exists c\in\mathcal{C}_t,\ \mathrm{match}(v_{t,k},c)=1\}\right|,
  \qquad
  \mathrm{RL}_s=\overline{\mathrm{RL}}_s .
  \label{eq:rl}
\end{equation}

\subsection{Utility Results}
\label{sec:utility}
Table~\ref{tab:main} shows that answer quality is set more by the form of evidence than by the volume of context.
With the Arch held federated, replacing raw passages with typed evidence lifts relevance, correctness, and nugget recall from the collapse of plain retrieval to a competitive level on all four models.
In the collapsed baseline, faithfulness stays high while relevance and recall fall.
The answers are grounded in retrieved text but fail to address the question's conditions.
Typed evidence separates extraction from composition, so facts are extracted once per domain at retrieval time and the generator composes compact records with stable sources, conditions, and missing facts during answer synthesis.

\begin{table}[t]
  \centering
  \caption{
    Utility and token-cost results on the multi-turn policy benchmark for the six frameworks.
    \textbf{Bold} marks the best value and \underline{underline} the second best in a metric.
  }
  \label{tab:main}
  \footnotesize
  \setlength{\tabcolsep}{5pt}
  \resizebox{1\textwidth}{!}{%
  \begin{tabular}{llll rrrrrrrr}
    \toprule
    Arch & Method & Model & Variant
      & CR~$\uparrow$ & Faith.~$\uparrow$ & Rel.~$\uparrow$ & Corr.~$\uparrow$
      & Acc.~$\uparrow$ & NR~$\uparrow$ & CO~$\uparrow$ & Tok/T \\
    \midrule
    \multirow{8}{*}{Cent}
      & \multirow{4}{*}{Agentic RAG} & \multirow{2}{*}{GPT} & 4.1-nano & 51.94 & 82.18 & 69.52 & 58.92 & 61.31 & 34.77 & 93.31 & 22.04 \\
      & & & 4.1-mini & 58.63 & 91.69 & 74.99 & 67.28 & 64.96 & 42.99 & 97.07 & 20.30 \\
      & & \multirow{2}{*}{Qwen} & 3.5-9B & 50.93 & 92.74 & 72.82 & 66.03 & 68.06 & 47.19 & 87.80 & 21.26 \\
      & & & 3.6-27B & 48.88 & 97.39 & 80.99 & 74.72 & 70.33 & 57.63 & 83.58 & 20.63 \\
    \cmidrule(l){2-12}
      & \multirow{4}{*}{\makecell[l]{\quad + History}} & \multirow{2}{*}{GPT} & 4.1-nano & 52.75 & 79.81 & 69.14 & 58.48 & 61.38 & 32.13 & 90.87 & 23.77 \\
      & & & 4.1-mini & 59.38 & 89.91 & 73.49 & 66.69 & 64.05 & 45.17 & 97.44 & 21.95 \\
      & & \multirow{2}{*}{Qwen} & 3.5-9B & 52.17 & 91.09 & 71.27 & 64.54 & 67.84 & 48.33 & 90.72 & 22.92 \\
      & & & 3.6-27B & 50.45 & 96.29 & 79.05 & 72.48 & 70.74 & 59.91 & 87.36 & 22.23 \\
    \midrule
    \multirow{16}{*}{Fed}
      & \multirow{4}{*}{RAG} & \multirow{2}{*}{GPT} & 4.1-nano & 40.03 & 77.72 & 27.44 & 22.65 & 61.30 & 4.69 & 93.47 & 9.30 \\
      & & & 4.1-mini & 41.53 & 81.09 & 32.73 & 29.26 & 61.30 & 7.80 & 92.91 & 9.35 \\
      & & \multirow{2}{*}{Qwen} & 3.5-9B & 41.16 & 82.23 & 29.16 & 25.63 & 61.31 & 3.96 & 90.90 & 10.05 \\
      & & & 3.6-27B & 43.46 & 83.26 & 32.10 & 28.22 & 61.30 & 5.57 & 87.46 & 9.89 \\
    \cmidrule(l){2-12}
      & \multirow{4}{*}{Agentic RAG} & \multirow{2}{*}{GPT} & 4.1-nano & 63.02 & 87.16 & 70.88 & 61.88 & 61.28 & 33.44 & 97.26 & 13.71 \\
      & & & 4.1-mini & 64.20 & 92.91 & 75.37 & 68.42 & 65.31 & 47.24 & 96.77 & 13.00 \\
      & & \multirow{2}{*}{Qwen} & 3.5-9B & 58.61 & 94.02 & 74.09 & 67.82 & 68.32 & 48.99 & 88.79 & 13.62 \\
      & & & 3.6-27B & 57.64 & 96.63 & 82.12 & 75.46 & 70.01 & 59.10 & 82.79 & 13.09 \\
    \cmidrule(l){2-12}
      & \multirow{4}{*}{FARAG} & \multirow{2}{*}{GPT} & 4.1-nano & 58.24 & 82.04 & 63.28 & 55.07 & 61.34 & 25.09 & \textbf{99.62} & 15.16 \\
      & & & 4.1-mini & \textbf{65.56} & 93.57 & 76.28 & 70.25 & 66.40 & 47.81 & 95.08 & 14.06 \\
      & & \multirow{2}{*}{Qwen} & 3.5-9B & 60.76 & 94.57 & 74.78 & 68.27 & 68.32 & 51.23 & 91.08 & 14.96 \\
      & & & 3.6-27B & 62.84 & \textbf{97.82} & \textbf{84.65} & \textbf{78.66} & \textbf{71.35} & \textbf{65.75} & 84.46 & 14.20 \\
    \cmidrule(l){2-12}
      & \multirow{4}{*}{\makecell[l]{\quad + TPT}} & \multirow{2}{*}{GPT} & 4.1-nano & 57.63 & 82.42 & 62.39 & 53.65 & 61.28 & 23.90 & \underline{99.06} & 15.13 \\
      & & & 4.1-mini & \underline{64.43} & 93.54 & 76.48 & 70.07 & 66.38 & 48.59 & 94.35 & 14.02 \\
      & & \multirow{2}{*}{Qwen} & 3.5-9B & 60.45 & 93.67 & 74.23 & 67.79 & 67.93 & 47.76 & 90.48 & 14.94 \\
      & & & 3.6-27B & 62.08 & \underline{97.56} & \underline{83.78} & \underline{77.62} & \underline{70.82} & \underline{64.39} & 84.56 & 14.21 \\
    \bottomrule
  \end{tabular}}
\end{table}

Adding query history on top of typed evidence has a capacity-dependent sign.
Within each model family, the gain from history grows with model size and turns negative on the smallest model.
Query history adds relevant conversational context along with irrelevant prior turns.
This moves the design question from whether to attach history to where it is exposed and how user attributes are released.
Citation Overlap and Accuracy require separate reading.
Citation Overlap penalizes short citation lists against exhaustive gold sources.
Accuracy is an answerability decision score, so it can remain stable even when content scores change across matched settings and model families.

\subsection{Privacy and Token-Cost Results}
Table~\ref{tab:privacy} shows that FARAG+TPT drives both raw-leakage counts to zero on every model.
Between plain FARAG and FARAG+TPT, domain pre-filtering and typed evidence are held fixed.
The release form is the only changed factor.
Answer quality in Table~\ref{tab:main} gives up only a small margin.
The result is consistent with the task structure: eligibility answers usually turn on conditions already carried by the evidence, so the attribute type supports synthesis while the exact value mainly increases exposure along the transmitted path.

\begin{table}[t]
  \centering
  \caption{
    Privacy on the policy benchmark for the three attribute-sharing conditions.
    \textbf{Bold} marks the best performance in a metric with the preferred direction.
    }
  \label{tab:privacy}
  \footnotesize
  \setlength{\tabcolsep}{6pt}
  \resizebox{.95\textwidth}{!}{%
  \begin{tabular}{llll rrrr}
    \toprule
    Arch & Method & Model & Variant
      & AKR & UER~$\downarrow$
      & SE~$\downarrow$ & RL~$\downarrow$ \\
    \midrule
    \multirow{4}{*}{Cent} & \multirow{4}{*}{\makecell[l]{Agentic RAG \\\quad + History}} & \multirow{2}{*}{GPT} & 4.1-nano & 0.9546 & 0.9626 & 5.1981 & 5.1981 \\
      & & & 4.1-mini & 0.9543 & 0.9625 & 5.1981 & 5.1981 \\
      & & \multirow{2}{*}{Qwen} & 3.5-9B & 0.9548 & 0.9624 & 5.1981 & 5.1981 \\
      & & & 3.6-27B & 0.9540 & 0.9625 & 5.1981 & 5.1981 \\
    \midrule
    \multirow{8}{*}{Fed} & \multirow{4}{*}{FARAG} & \multirow{2}{*}{GPT} & 4.1-nano & 0.7626 & 0.3490 & 4.3306 & 1.7147 \\
      & & & 4.1-mini & 0.5790 & 0.1989 & 1.6014 & 1.0261 \\
      & & \multirow{2}{*}{Qwen} & 3.5-9B & 0.6612 & 0.2930 & 2.3771 & 1.4858 \\
      & & & 3.6-27B & 0.3990 & 0.1271 & 0.8902 & 0.6746 \\
    \cmidrule(l){2-8}
      & \multirow{4}{*}{\makecell[l]{\quad + TPT}} & \multirow{2}{*}{GPT} & 4.1-nano & 0.7612 & 0.3442 & \textbf{0.0000} & \textbf{0.0000} \\
      & & & 4.1-mini & 0.5771 & 0.1996 & \textbf{0.0000} & \textbf{0.0000} \\
      & & \multirow{2}{*}{Qwen} & 3.5-9B & 0.6599 & 0.2896 & \textbf{0.0000} & \textbf{0.0000} \\
      & & & 3.6-27B & 0.3897 & \textbf{0.1252} & \textbf{0.0000} & \textbf{0.0000} \\
    \bottomrule
  \end{tabular}}
\end{table}

Domain scoping also lowers token cost.
In Table~\ref{tab:main}, per-turn cost falls on every model when the Arch moves from centralized to federated at a matched method.
Within the federated block, typed evidence costs more tokens than plain retrieval because domain agents read passages to write it.
That spending buys the quality gains in Table~\ref{tab:main}.
Sending types in place of values adds no measured token overhead, so FARAG+TPT matches the token cost of plain FARAG.

\subsection{On-device Deployment}
The federated split is also a deployment claim: FARAG keeps the persona agent on the user side and allows domain agents to run either serially on one local device or in parallel across domain devices or services.
We measure per-turn deployment cost for the locally deployable Qwen models on one NVIDIA RTX 4090 with 4-bit \texttt{Q4\_K\_M} weights under \texttt{llama.cpp}.
Table~\ref{tab:deploy} reports Calls, the average number of model invocations per turn; Prompt, input prompt tokens in thousands; Comm, inter-agent communication in kilobytes; Seq, single-device sequential latency; Par, latency when domain-agent calls are placed in parallel; and VRAM and Disk, the peak resident memory and model footprint.

\begin{table}[t]
  \centering
  \caption{
    Per-turn on-device deployment cost of Qwen models.
  }
  \label{tab:deploy}
  \scriptsize
  \setlength{\tabcolsep}{4pt}
  \resizebox{\textwidth}{!}{%
  \begin{tabular}{lll rr r rr rr}
    \toprule
    Arch & Method & Model & Calls & Prompt & Comm & Seq & Par & VRAM & Disk \\
     & & & /turn & kTok & KB & (s) & (s) & (GB) & (GB) \\
    \midrule
    \multirow{4}{*}{Cent}
      & \multirow{2}{*}{Agentic RAG} & 3.5-9B & 5.9 & 20.2 & 12.2 & 16.6 & 6.4 & 8.5 & 5.5 \\
      & & 3.6-27B & 5.8 & 19.8 & 10.9 & 29.2 & 11.5 & 22.0 & 16.5 \\
    \cmidrule(l){2-10}
      & \multirow{2}{*}{\makecell[l]{\quad + History}} & 3.5-9B & 5.9 & 21.7 & 15.2 & 18.2 & 7.1 & 8.5 & 5.5 \\
      & & 3.6-27B & 5.8 & 21.3 & 13.8 & 32.0 & 12.9 & 22.0 & 16.5 \\
    \midrule
    \multirow{8}{*}{Fed}
      & \multirow{2}{*}{RAG} & 3.5-9B & 1.0 & 9.7 & 44.4 & 6.4 & 6.4 & 8.5 & 5.5 \\
      & & 3.6-27B & 1.0 & 9.5 & 44.4 & 12.7 & 12.7 & 22.0 & 16.5 \\
    \cmidrule(l){2-10}
      & \multirow{2}{*}{Agentic RAG} & 3.5-9B & 3.9 & 12.7 & 8.6 & 13.1 & 6.8 & 8.5 & 5.5 \\
      & & 3.6-27B & 3.8 & 12.4 & 7.4 & 22.6 & 12.3 & 22.0 & 16.5 \\
    \cmidrule(l){2-10}
      & \multirow{2}{*}{FARAG} & 3.5-9B & 3.9 & 13.9 & 9.0 & 15.3 & 7.9 & 8.5 & 5.5 \\
      & & 3.6-27B & 3.8 & 13.3 & 7.7 & 26.3 & 13.2 & 22.0 & 16.5 \\
    \cmidrule(l){2-10}
      & \multirow{2}{*}{\makecell[l]{\quad + TPT}} & 3.5-9B & 3.9 & 13.8 & 8.9 & 15.3 & 7.8 & 8.5 & 5.5 \\
      & & 3.6-27B & 3.8 & 13.3 & 7.6 & 26.4 & 13.2 & 22.0 & 16.5 \\
    \bottomrule
  \end{tabular}}
\end{table}

Table~\ref{tab:deploy} shows the same mechanism as the token-cost results in Table~\ref{tab:main}: domain routing removes off-domain reading before evidence construction, so the history-bearing FARAG variants reduce prompt volume and sequential latency relative to the centralized history-bearing setting.
Plain federated RAG is the cheapest row because it issues a single retrieval-to-answer call, yet Table~\ref{tab:main} shows that this cheap path collapses answer quality; typed evidence spends additional calls to turn retrieved passages into usable domain records.
Typed release changes the representation of transmitted attributes rather than the call graph, so FARAG+TPT tracks plain FARAG in deployment cost while eliminating raw sensitive-value leakage.
The gap between Seq and Par also illustrates the deployment value of the federation: domain-agent calls are separable and can be overlapped across devices or services, whereas the centralized settings retain a single shared context path.
Finally, memory footprint is dominated by model size, making the smaller Qwen model the comfortable on-device target and the larger model feasible but memory-bound on a consumer GPU.

\section{Related Work}
\label{sec:related}

\paragraph{Retrieval and agentic RAG.}
RAG grounds language-model outputs in retrieved text \cite{lewis2020rag,karpukhin2020dpr,gao2024ragsurvey}, while dense and multi-hop retrievers extend grounding to evidence chains \cite{xiong2021multihop,khattab2021baleen}.
Conditional QA and legal RAG benchmarks show that rule-governed answers depend on both document evidence and user-specific conditions \cite{sun2022conditionalqa,pipitone2024legalbenchrag}.
Irrelevant passages can still distract the generator \cite{yoran2024robust,amiraz2025distracting}.
Agentic RAG separates retrieval, reasoning, and tool use \cite{yao2023react,asai2024selfrag}, but stepwise control alone does not specify which corpus, evidence form, or user-attribute field each step may read.
FARAG keeps grounding while making corpus, evidence, query-history visibility, and attribute release explicit within the same pipeline for every turn.

\paragraph{Privacy, memory, and edge-cloud agents.}
Contextual integrity restricts information flow to licensed parties \cite{nissenbaum2004privacy}, and AirGapAgent applies this idea to assistant memory through task-based filtering \cite{bagdasarian2024airgap}.
RAG privacy studies, memory-agent benchmarks, and edge-cloud agent systems show that private corpora, persistent user attributes, and placement choices become disclosure surfaces across turns \cite{zeng2024ragprivacy,wang2025memoryprivacy,packer2024memgpt,maharana2024locomo,pan2025secom,wang2025slmllmsurvey,yu2026plantwin,yuan2026paac,chen2026memprivacy,zhan2026prism}.
FARAG applies this boundary view by coupling privacy control to evidence allocation during retrieval, release, and final synthesis.

\paragraph{Evaluation.}
LLM judges score generated answers through rubric-based evaluation \cite{zheng2023judging,liu2023geval}, RAGAS separates relevance and faithfulness \cite{es2024ragas}, and nugget-based RAG evaluation measures recovery of passage-grounded required facts \cite{pradeep2025greatnuggetrecall}.
We therefore report answer quality, structural privacy, and token cost on the benchmark turns used by the experimental tables in this section and all reported runs.

\section{Conclusion}
\label{sec:conclusion}
FARAG allocates context across query history, user attributes, and domain evidence with a simple rule: full attributes are read on the user side, and the persona agent controls each release.
Across four models, typed evidence drives quality, domain scoping reduces cost, and typed release removes raw leakage.
Future work will test real consultations, bound leakage from released types, and assess whether cross-domain evidence improves the quality of policy advice.

\bibliographystyle{splncs04}
\bibliography{references}

\end{document}